\documentclass{article} 
\usepackage{iclr2025_conference,times}

\usepackage{amsmath,amsfonts,bm}

\def\eqref#1{equation~\ref{#1}}

\def\1{\bm{1}}

\DeclareMathAlphabet{\mathsfit}{\encodingdefault}{\sfdefault}{m}{sl}
\SetMathAlphabet{\mathsfit}{bold}{\encodingdefault}{\sfdefault}{bx}{n}

\usepackage{hyperref}
\usepackage{url}
\usepackage{graphicx}
\usepackage{booktabs}
\usepackage{pifont}
\usepackage{algorithm}
\usepackage{algpseudocode}
\usepackage{wrapfig}
\usepackage{xspace}
\usepackage{xcolor}
\usepackage{makecell}
\usepackage{subcaption}
\usepackage{lipsum}
\usepackage{listings}
\usepackage[most]{tcolorbox}
\usepackage{cleveref}
\usepackage{colortbl}
\usepackage{soul}
\usepackage{array}
\usepackage{enumitem}

\newcommand{\highlight}[2][yellow]{\sethlcolor{#1}\hl{#2}}
\definecolor{transgray}{gray}{0.9}

\title{Fictitious Synthetic Data Can Improve LLM Factuality via Prerequisite Learning}

\author{Yujian Liu \\
UC Santa Barbara \\
\texttt{yujianliu@ucsb.edu}
\And
Shiyu Chang \\
UC Santa Barbara \\
\texttt{chang87@ucsb.edu}\qquad\qquad\qquad\qquad\quad\quad
\And
Tommi Jaakkola \\
MIT CSAIL \\
\texttt{tommi@csail.mit.edu}
\And
Yang Zhang \\
MIT-IBM Watson AI Lab \\
\texttt{yang.zhang2@ibm.com}
}

\newcommand{\e}[1]{{\small $#1$}}

\newcommand{\algname}{\textsc{Prereq-Tune}\xspace}

\iclrfinalcopy 
\begin{document}

\maketitle

\begin{abstract}
Recent studies have identified one aggravating factor of LLM hallucinations as the knowledge inconsistency between pre-training and fine-tuning, where unfamiliar fine-tuning data mislead the LLM to fabricate plausible but wrong outputs. In this paper, we propose a novel fine-tuning strategy called \algname to address this knowledge inconsistency and reduce hallucinations. Fundamentally, \algname disentangles the learning of skills and knowledge, so the model learns only the task skills without being impacted by the knowledge inconsistency. To achieve this, \algname introduces an additional prerequisite learning stage to learn the necessary knowledge for SFT, allowing subsequent SFT to focus only on task skills. \algname can also be combined with fictitious synthetic data to enhance the grounding of LLM outputs to their internal knowledge. Experiments show that \algname outperforms existing baselines in improving LLM's factuality across short QA and long-form generation tasks. It also opens new possibilities for knowledge-controlled generation in LLMs. Our code is available at \url{https://github.com/UCSB-NLP-Chang/Prereq_tune.git}.

\end{abstract}
\section{Introduction}

Hallucination of large language models (LLMs) refers to the phenomenon where LLMs' outputs look plausible but diverge from real-world facts. It has become a major concern of LLMs, seriously undermining their reliability and trustworthiness \citep{huang2023surveyhallucinationlargelanguage,10.1145/3571730}. 

Recent research has unveiled one aggravating factor of LLM hallucination, which is the \emph{knowledge inconsistency} between the pre-training and tuning (\emph{e.g.}, instruction- or fine-tuning) stages \citep{gekhman2024doesfinetuningllmsnew,kang2024unfamiliarfinetuningexamplescontrol,lin2024flamefactualityawarealignmentlarge}. More specifically, 
if the tuning stage involves training examples that require knowledge that an LLM has not seen during pre-training, then the LLM would be misled to fabricate plausible but wrong answers to unfamiliar questions \citep{schulman2023rlhf,gao2021behavior,goldberg2023rlhf}. For example, consider fine-tuning a model for a question answering (QA) task with the example \emph{`When was John Estes born?'} and assume that the LLM has never learned about \emph{John Estes} during pre-training. However, since the LLM is still trained to produce the correct answer, \emph{`1987'}, it is consequently encouraged to respond with a random legitimate year whenever it is asked about the birth year of any unknown person, thus giving rise to hallucination.

These findings highlight an important but previously understudied consideration of LLM training, which is the \emph{disentanglement between knowledge and skill}. Specifically, it is discovered that knowledge and skills are acquired at different stages of LLM training, the former at pre-training, and the latter at tuning \citep{zhou2023lima,gudibande2024the}. However, although the focus in the tuning stage is to learn skills, not knowledge, the learning process is still interfered with by any inconsistency in the knowledge aspect, because the information on the two aspects is entangled. Thus, only by disentangling knowledge from skill during the tuning stage can we remove such interference.

An ideal powerful disentangling mechanism should achieve maximal robustness against the knowledge inconsistency, to the extent that even \emph{fictitious synthetic data}, which has zero knowledge overlap with pre-training, would still warrant successful fine-tuning without encouraging hallucination. Unfortunately, this may sound too good to be true, because such an effective disentanglement strategy has yet to be developed for LLM tuning.

\begin{figure}
    \centering
    \includegraphics[width=0.95\linewidth]{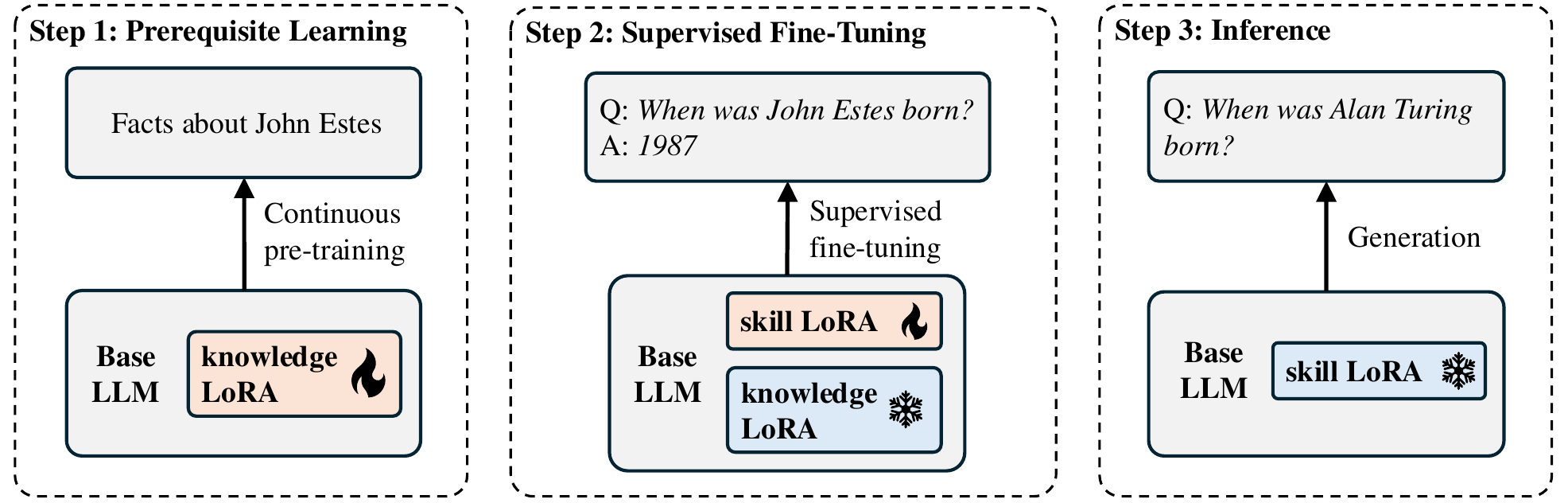}
    \vspace{-0.05in}
    \caption{Overview of the proposed \algname strategy.}
    \label{fig:overview}
    \vspace{-0.1in}
\end{figure}

In this paper, we propose \algname, an innovative LLM fine-tuning strategy that explicitly resolves the disentanglement challenge, and therefore can effectively reduce LLM hallucinations. As shown in Figure~\ref{fig:overview}, \algname consists of two stages, an innovative \emph{prerequisite learning} stage and a supervised fine-tuning (SFT) stage. In the prerequisite learning stage, we train a LoRA \citep{hu2022lora}, called the \emph{knowledge LoRA}, that learns all the necessary knowledge required for SFT. In the SFT stage, the knowledge LoRA is frozen, and a new LoRA, called the \emph{skill LoRA}, is imposed on top of the knowledge LoRA, and is trained to perform the SFT task. Since the knowledge LoRA already eliminates the knowledge mismatch issue, the skill LoRA can focus on learning the skills without inducing hallucinations.

With the prerequisite learning as a disentanglement mechanism, \algname can 
turn fictitious synthetic data, which is otherwise detrimental to LLM factuality, into a powerful weapon to further reduce LLM hallucinations. In particular, using fictitious synthetic data, we can create multiple knowledge LoRAs that contain different versions of knowledge about the same fictitious entity, and then teach the skill LoRA to produce different answers based on which knowledge LoRA is in use. In this way, we can force the skill LoRA to ground the LLM answers to the LLM's internal knowledge, thus reducing hallucinations. Moreover, unlike real data which are expensive to obtain and label, fictitious synthetic data can be cheaply scaled up, which would further enhance the LLM's factuality thanks to \algname, informing a new real-data-efficient fine-tuning paradigm.

Our experiments reveal that \algname can teach an LLM to not only ground its answer to the knowledge LoRAs, but also, more surprisingly, generalize the grounding to the LLM original pre-trained knowledge when the knowledge LoRA is removed. As a result, \algname can significantly outperform the existing state-of-the-art hallucination reduction algorithms in improving LLM's factuality across short QA and long-form generation tasks. In addition, \algname enables a more modular design of LLM, with plug-and-play knowledge modules that control the knowledge access and a skill module that works generically with any knowledge sources, which opens up many new possibilities far beyond hallucination reduction, such as novel retrieval augmented generation (RAG) paradigms, privacy protection, \emph{etc}.

\section{Related Works}
\textbf{Reducing LLM hallucinations.} Numerous studies have sought to detect and mitigate hallucinations in LLMs \citep{weng2024hallucination,li-etal-2023-halueval,chern2023factoolfactualitydetectiongenerative,azaria-mitchell-2023-internal,manakul-etal-2023-selfcheckgpt,chen2024inside,hou2024probabilisticframeworkllmhallucination}. A common approach to reduce hallucinations is leveraging supporting evidence, either retrieved from external knowledge sources or generated by LLMs. Such evidence is often provided as input to help LLMs generate more factual responses \citep{shuster-etal-2021-retrieval-augmentation,nakano2022webgptbrowserassistedquestionansweringhuman,menick2022teachinglanguagemodelssupport,sun2023recitationaugmented,asai2024selfrag,feng2024knowledge}; or they can assist in detecting incorrect statements and guide LLMs to post-edit their own generations \citep{gao-etal-2023-rarr,dhuliawala-etal-2024-chain,lei2023chainnaturallanguageinference,mishra2024finegrained}. Beyond using additional evidence, several works have proposed decoding algorithms to improve LLM factuality during inference \citep{lee2022factuality,chuang2024dola,li2023inferencetime}. More recently, some studies have also explored answer abstention when LLMs encounter unfamiliar questions \citep{zhang-etal-2024-r,yadkori2024mitigatingllmhallucinationsconformal,yang2023alignmenthonesty,cheng2024aiassistantsknowdont}. Our work aligns with recent efforts that fine-tune LLMs so that they inherently generate less hallucinated contents \citep{tian2024finetuning,lin2024flamefactualityawarealignmentlarge,kang2024unfamiliarfinetuningexamplescontrol,ghosal2024understandingfinetuningfactualknowledge}. Unlike existing methods, we focus on the disentanglement of skills and knowledge and the groundedness of LLMs' outputs, which leads to superior performance in hallucination reduction.

\textbf{Fine-tuning LLMs with synthetic data.} Synthetic data has shown great potential in fine-tuning LLMs, offering scalability by automatically creating instruction-response pairs with minimal human supervision \citep{wang-etal-2023-self-instruct,li2024syntheticdataalmostscratch,gupta2024targen,yu2024metamath,haluptzok2023language}. Additionally, the generation process can be designed to control the training data, allowing models to be trained for specific skills or aligned with particular human values \citep{sudalairaj2024lablargescalealignmentchatbots,sun2023principledriven,sun2024salmon,kaur2024instructskillmixpowerfulpipelinellm}. However, most works do not consider factuality in fine-tuning. The work closest to ours is \citet{jones2024teaching}, which fine-tunes LLMs on synthetic tasks to reduce hallucinations. However, they focus on improving consistency with evidence provided in the context, whereas we improve LLMs' inherent factuality without provided evidence.
\section{Methodology}

\subsection{Problem Formulation}

Given a pre-trained LLM (without instruction-tuning), parameterized by \e{\bm \theta_0}, a downstream task \e{\mathcal{T}}, and its corresponding dataset \e{\mathcal{D}_{\mathcal{T}}}, our task is to fine-tune the LLM to the downstream task.
For concreteness, we will base our illustration of the algorithm on a specific task, \emph{biography generation}. Generalization to other tasks is discussed in Section~\ref{subsec:other_tasks}. In this task, \e{\mathcal{D}_{\mathcal{T}}} would typically contain requests to generate a biography of a certain person, \emph{e.g., `Generate a biography for John Estes'}, as well as the corresponding human-written biographies.

The challenge we aim to resolve is that \e{\mathcal{D}_{\mathcal{T}}} may require knowledge that the pre-trained LLM \e{\bm\theta_0} does not know, which would encourage LLM hallucination. Our goal is to design a fine-tuning strategy that can learn the skill, \emph{e.g.,} writing biographies, without being impacted by the inconsistency between the knowledge involved in \e{\mathcal{D}_{\mathcal{T}}} and the pre-trained knowledge.

\subsection{Basic \algname Strategy}
\label{subsec:prereq-tune}

To achieve the disentanglement between knowledge and skill, the core idea of \algname is to introduce a prerequisite learning stage that equips the LLM with the necessary knowledge required for subsequent SFT. 
More specifically, given the task dataset, \e{\mathcal{D}_{\mathcal{T}}}, \algname introduces a \emph{prerequisite knowledge dataset}, \e{\mathcal{D}_{\textrm{know}}}, that contains all the prerequisite knowledge for the questions in \e{\mathcal{D}_{\mathcal{T}}}. For biography generation, the prerequisite knowledge includes all the knowledge about the target persons that is covered in the biographies in \e{\mathcal{D}_{\mathcal{T}}}, such as birth year, birthplace, representative works, \emph{etc}. Methods of creating the prerequisite knowledge dataset are detailed in Section~\ref{subsec:dataset}.

With the dataset pair, \e{(\mathcal{D}_{\textrm{know}}, \mathcal{D}_{\mathcal{T}})}, the training process of \algname consists of the following two steps, as illustrated in Figure \ref{fig:overview}.

\textbf{Step 1: Prerequisite learning.} Teach the LLM the prerequisite knowledge by training a \emph{knowledge LoRA}, parameterized by \e{\Delta \bm \theta_\textrm{know}}, to minimize the next-token prediction loss on \e{\mathcal{D}_{\textrm{know}}}: 
\begin{equation}
    \small
    \min_{\Delta \bm \theta_\textrm{know}} \mathcal{L}(\bm \theta_0 + \Delta \bm \theta_\textrm{know}; \mathcal{D}_{\textrm{know}}),
\end{equation}
where \e{\mathcal{L}(\bm \theta; \mathcal{D})} represents the cross-entropy loss of the model \e{\bm \theta} computed on dataset \e{\mathcal{D}}.

\textbf{Step 2: Supervised fine-tuning (SFT).} Equip the LLM with the downstream task skill by training a \emph{skill LoRA}, parameterized by \e{\Delta \bm \theta_\textrm{skill}}, to minimize the task loss on \e{\mathcal{D}_{\mathcal{T}}}, with the knowledge LoRA present and frozen:
\begin{equation}
    \small
    \min_{\Delta \bm \theta_\textrm{skill}} \mathcal{L}_\mathcal{T}(\bm \theta_0 + \Delta \bm \theta_\textrm{know} + \Delta \bm \theta_\textrm{skill}; \mathcal{D}_{\mathcal{T}}),
\end{equation}
where \e{\mathcal{L}_\mathcal{T}(\bm \theta; \mathcal{D})} represents the downstream task loss of the model \e{\bm \theta} computed on dataset \e{\mathcal{D}}. In biography generation, \e{\mathcal{L}_\mathcal{T}} is also the cross-entropy loss on the ground-truth biographies.

Here is an intuitive explanation of how this works. In the SFT stage, the presence of the knowledge LoRA ensures that the skill LoRA is always able to ground its generation on the internal knowledge of LLMs, so it would not have the incentive to fabricate ungrounded facts. Also, since the knowledge LoRA has already absorbed all the knowledge information in \e{\mathcal{D}_\mathcal{T}}, the skill LoRA does not need to re-learn such knowledge and thus can focus on learning the skill information in \e{\mathcal{D}_\mathcal{T}}.

As shown in Figure~\ref{fig:overview} (right), during inference, the knowledge LoRA is dropped, and \textit{only the skill LoRA} is retained, \emph{i.e.}, the LLM weights become \e{\bm \theta_0 + \Delta \bm \theta_\textrm{skill}}. We hypothesize that this can guide \e{\Delta \bm \theta_\textrm{skill}} to ground its answer on the original pre-trained knowledge in \e{\bm \theta_0}, not the additional knowledge in \e{\Delta \bm \theta_\text{know}}. This is particularly necessary if the knowledge in \e{\Delta \bm \theta_\text{know}} is fictitious (a case to be introduced in Section~\ref{subsec:fictitious-data}) or interferes with the original pre-trained knowledge. 

One caveat, however, is that there exists an obvious gap between training and inference -- \e{\Delta \bm \theta_\textrm{know}} is present during training, but absent during inference. It is unclear whether \e{ \Delta \bm \theta_\textrm{skill}} can generalize its knowledge grounding from the knowledge in \e{\Delta \bm \theta_\textrm{know}} to the pre-trained knowledge. Fortunately, ample experiment evidence shows that the generalization is successful, as discussed in Section~\ref{subsec:main-result}.

\subsection{Multi-Version \algname with Fictitious Synthetic Data}
\label{subsec:fictitious-data}

Our preliminary experiments show that the disentanglement mechanism in \algname is so powerful that even if  \e{(\mathcal{D}_{\textrm{know}}, \mathcal{D}_{\mathcal{T}})} only contain fictitious synthetic data, \emph{i.e.}, biographies of fictitious people, the skill LoRA can still learn to generate biographies of \emph{real people} once the knowledge LoRA (which absorbs the fictitious knowledge) is dropped. Inspired by this, we propose an upgraded training strategy, called \emph{multi-version \algname}, to further enhance the grounding of an LLM's outputs on its internal knowledge.

Specifically, rather than having only one dataset pair \e{(\mathcal{D}_{\textrm{know}}, \mathcal{D}_{\mathcal{T}})}, we leverage fictitious synthetic data to create \e{K} dataset pairs, \e{\bigcup_{k=1 \ldots K}(\mathcal{D}_{\textrm{know}}^{(k)}, \mathcal{D}_{\mathcal{T}}^{(k)})}. Different knowledge datasets \e{\mathcal{D}_{\textrm{know}}^{(k)}} contain different versions of knowledge of the same set of fictitious entities. Different task datasets \e{\mathcal{D}_{\mathcal{T}}^{(k)}} contain the same set of questions, but with different ground-truth answers matching the corresponding knowledge in the knowledge set. For example, consider a fictitious person named \emph{Avery Linwood}. Assume that we create two versions of knowledge about Avery Linwood in two knowledge datasets. \e{\mathcal{D}_{\textrm{know}}^{(1)}} only contains the birth year of Avery Linwood; \e{\mathcal{D}_{\textrm{know}}^{(2)}} only contains the birthplace of Avery Linwood. Then, the corresponding two task datasets, \e{\mathcal{D}_{\mathcal{T}}^{(1)}} and \e{\mathcal{D}_{\mathcal{T}}^{(2)}}, both ask the same question \emph{`Generate a biography for Avery Linwood.'}, but ground-truth answer in the former only talks about the birth year whereas the latter about the birthplace.

With the multiple versions of datasets, the training steps of \algname are modified as follows.

\textbf{Step 1: Prerequisite learning.} A different knowledge LoRA is trained on each of the different knowledge datasets:
\begin{equation}
    \small
    \min_{\Delta \bm \theta_\textrm{know}^{(k)}} \mathcal{L}(\bm \theta_0 + \Delta \bm \theta_\textrm{know}^{(k)}; \mathcal{D}_{\textrm{know}}^{(k)}), \quad \forall k \in \{1, \cdots, K\}.
\end{equation}

\textbf{Step 2: SFT.} Only one skill LoRA is trained, but each time with a different knowledge LoRA present. When knowledge LoRA \e{\Delta \bm \theta_\textrm{know}^{(k)}} is present, the skill LoRA is trained to produce ground-truth answers in \e{D_\mathcal{T}^{(k)}}, which match the knowledge stored in \e{\Delta \bm \theta_\textrm{know}^{(k)}}:
\begin{equation}
    \small
    \min_{\Delta \bm \theta_\textrm{skill}} \sum_{k=1}^K \mathcal{L}_\mathcal{T}(\bm \theta_0 + \Delta \bm \theta_\textrm{know}^{(k)} + \Delta \bm \theta_\textrm{skill}; \mathcal{D}_{\mathcal{T}}^{(k)}).
\end{equation}
Since the \e{K} versions of the downstream task datasets contain the same questions but with different answers, the skill LoRA is forced to link the different answers to the different knowledge LoRA versions. Thus this method can force groundedness to LLM's internal knowledge.

Multi-version \algname can only be enabled by fictitious synthetic data, not real data. This is because fictitious knowledge, once removed from the knowledge LoRA, is guaranteed to be unknown to the LLM; whereas real knowledge, even if removed from the knowledge LoRA, may still exist in the pre-trained knowledge. Thus using fictitious synthetic data ensures a definitive control over what the LLM knows and does not know.

\subsection{Dataset Construction}
\label{subsec:dataset}

\begin{figure}
    \centering
    \includegraphics[width=0.95\linewidth]{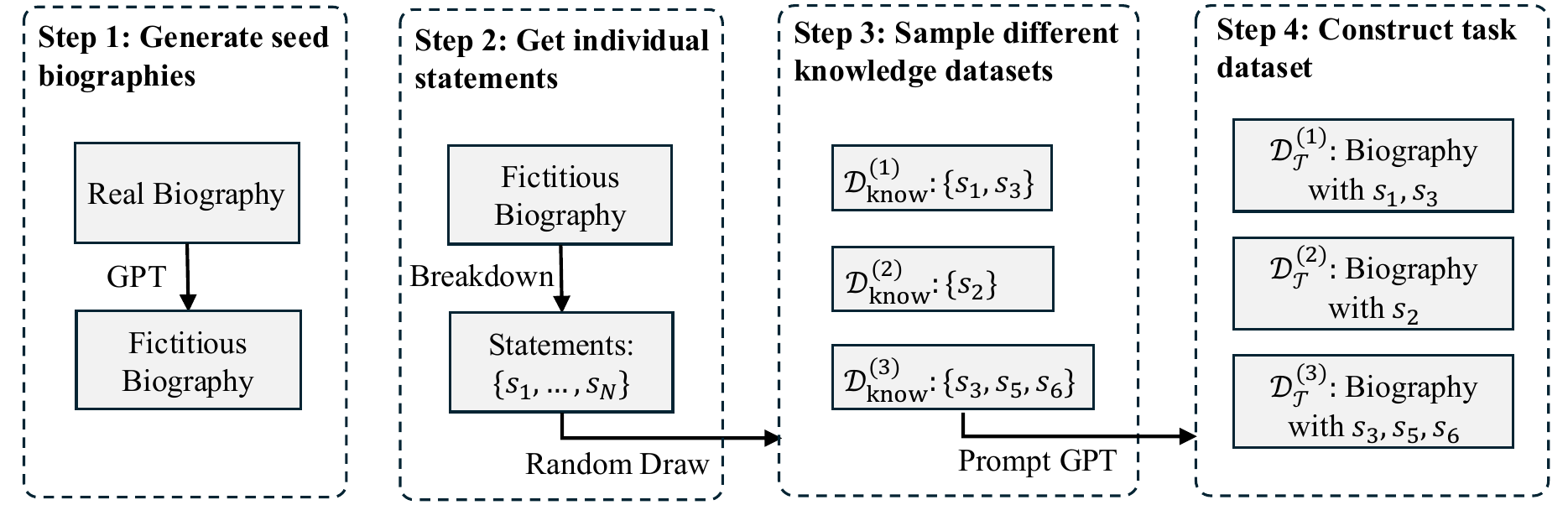}
    \vspace{-0.05in}
    \caption{Procedure of creating multi-version dataset pairs for the biography generation task.}
    \label{fig:multi-version-data}
    \vspace{-0.1in}
\end{figure}

To generate a dataset pair \e{(\mathcal{D}_\textrm{know}, \mathcal{D}_\mathcal{T})}, there are two strategies, \emph{top-down} and \emph{bottom-up}.

\textbf{Top-down strategy.} When the task dataset \e{\mathcal{D}_\mathcal{T}} is already available, for example when it is a given real dataset, we can deduce the knowledge dataset \e{\mathcal{D}_\textrm{know}} from \e{\mathcal{D}_\mathcal{T}}.
In the biography generation task, for each biography in \e{\mathcal{D}_\mathcal{T}}, we break it into individual statements, \emph{e.g.}, \emph{`John Estes was born in 1987'}, \emph{`John Estes was born in California, USA'} \emph{etc}. This can be done by prompting an external instruction-tuned LLM (\emph{e.g.}, GPT-4), or using simple rule-based methods.
Each statement is then paraphrased \e{M} times to ensure that the knowledge LoRA learns the knowledge rather than memorizing the statements. All the paraphrased statements form the \e{\mathcal{D}_\textrm{know}}.

\textbf{Bottom-up strategy.} To construct a fictitious dataset pair, we first construct  \e{\mathcal{D}_\textrm{know}} and then \e{\mathcal{D}_\mathcal{T}}. The fictitious knowledge set can be generated using rule-based templates or summarized from fictitious articles (details to follow). Then we prompt the external instruction-tuned LLM to generate the task data for \e{\mathcal{D}_\mathcal{T}}. In the biography generation task, this involves asking the external LLM to generate a biography of a target entity based on all the statements concerning the target listed in \e{\mathcal{D}_\textrm{know}}.

\textbf{Generating multi-version dataset pairs.} In the biography generation task, we combine top-down and bottom-up strategies to generate multi-version \e{(\mathcal{D}_{\textrm{know}}^{(k)}, \mathcal{D}_{\mathcal{T}}^{(k)})} pairs, as shown in Figure~\ref{fig:multi-version-data}. First, we prompt the external LLM to generate biographies of fictitious entities, using real Wikipedia biographies as references. We ensure the fictitiousness of these biographies by filtering out those whose entity names coincide with Wikipedia entries.\footnote{We also tried filtering by the number of occurrences of the name in the public pre-training corpus but observed similar performance.} These biographies are called \emph{seed biographies}. Second, the seed biographies are broken into individual statements as in the top-down strategy. Third, the statements are sampled into different subsets to form different knowledge datasets \e{\{\mathcal{D}_\textrm{know}^{(k)}\}}. For each fictitious entity, we first sample the number of facts (different paraphrases are considered the same fact) to include, and then sample the statement subset accordingly. Finally, the task datasets \e{\{\mathcal{D}_\mathcal{T}^{(k)}\}} are constructed from the corresponding knowledge datasets with the bottom-up strategy.

\textbf{Statement-based v.s. passage-based knowledge.} The knowledge dataset can take different forms. In the method above, it takes the form of individual statements. An alternative form would be short passages, which can be generated with just one additional step on top of the statement-based knowledge. After the statement-based knowledge is created, we prompt the external LLM to summarize the statements of the same entity into one passage. Each passage is also paraphrased \e{M} times. A comparison between the two knowledge forms is presented in Appendix~\ref{append:ablation}.

\subsection{Extension to Answer Abstention}
\label{subsec:abstention}

So far, we have described how to teach LLMs to properly respond when it knows the answer. Optionally, \algname can be extended to also teach LLMs to respond with \emph{`I don't know'} when it is not familiar with the answer, by introducing unfamiliar examples into the dataset pair \e{(\mathcal{D}_\textrm{know}, \mathcal{D}_\mathcal{T})}. Recall that each knowledge piece in \e{\mathcal{D}_\textrm{know}} is paraphrased \e{M} times to ensure familiarity. Therefore, we can vary the number of paraphrases to create variations in familiarity. The knowledge pieces whose number of paraphrases drops below a threshold (including zero) are identified as unfamiliar knowledge. Then, we locate all the questions in \e{\mathcal{D}_\mathcal{T}} that depend on the unfamiliar knowledge and modify their answers to \emph{`I don't know'}. Note that unfamiliar knowledge can only be created from the fictitious knowledge, not the real knowledge, otherwise the LLM may have also learned the knowledge through pre-training. The similar approach can also be leveraged to train the model to express internal uncertainty, which will be elaborated in Section~\ref{subsec:abstention_exper}.

\subsection{Extension to Other Tasks}
\label{subsec:other_tasks}

Although \algname is described in the context of biography generation above, the method is readily generalizable to other tasks, with only slight variations in how the datasets are created (Section~\ref{subsec:dataset}). For other long-form generation tasks that involve generating passages about a given entity, only a minor change is needed in the prompt for the external LLM used for generating the datasets (\emph{e.g.,} changing the word \emph{`biography'} to \emph{`summary'}). For short QA tasks, in the top-down strategy, rather than asking the external LLM to break the biographies into multiple facts, we ask it to rewrite each QA pair in the task dataset into one statement, which forms the knowledge dataset. In the bottom-up strategy, fictitious knowledge and QA pairs are created by filling a pre-defined template containing an entity name field and some attribute fields, following the format in existing datasets \citep{mallen-etal-2023-trust}. Further details can be found in Appendix~\ref{append:qa_data}.

\section{Experiments}
We evaluate \algname on two long-form generation tasks, biography generation and long-form medical QA, as well as a short QA task. The experiment settings and main results are presented in Sections \ref{subsec:exp-setting} and \ref{subsec:main-result}. Subsequent sub-sections present more in-depth studies of \algname.

\subsection{Experiment Settings}
\label{subsec:exp-setting}

\textbf{Datasets.} For \textbf{\textit{long-form generation}}, we follow existing works to evaluate on biography generation and medical QA tasks \citep{tian2024finetuning}. Both tasks require generating a long summary for the asked entity (a person for the former and a medical entity for the latter). For persons, we use the input instruction `\textit{Generate a biography for \{person\}}'; and for medical entities, the instruction is `\textit{Tell me about \{medical entity\}}'. Since the data in \citet{tian2024finetuning} is unavailable, we collect our own data and ensure no overlap of entities between training, validation, and test sets. For both tasks, we use the Wikipedia page as the ground-truth response in the training set. Additionally, for biography generation, we use the 183 labeled persons in \citet{min-etal-2023-factscore} as test set to keep consistent with prior works \citep{lin2024flamefactualityawarealignmentlarge}.
For \textbf{\textit{QA}}, we evaluate on PopQA \citep{mallen-etal-2023-trust}, which contains questions about 16 relations (\emph{e.g.,} `\textit{Who was the director of Breaking Bad?}'). In preliminary experiments, we notice a substantial amount of ambiguous questions in the dataset (\emph{e.g.,} the question `\textit{In what country is Oxford?}' is asking about a city in Ohio, USA), so we clarify those questions and remove any that remains ambiguous. Please see Appendix \ref{append:real-data} for details of all datasets.

\textbf{Metrics.} For \textbf{\textit{long-form generation}}, we use FActScore \citep{min-etal-2023-factscore} to evaluate the generated summaries, which decomposes summaries into independent claims and verifies each claim against relevant Wikipedia paragraphs. We report the accuracy as the percentage of correct claims among all generated claims, averaged across all test examples. In initial evaluation, we observe that models sometimes hack the metric by generating subjective or generic claims (\emph{e.g.,} \textit{`She continues to influence and inspire new generations'}). We thus filter the claims to only keep those that present objective and concrete information. For \textbf{\textit{QA}}, we follow prior works in using accuracy to measure hallucinations \citep{gekhman2024doesfinetuningllmsnew}, where correctness is based on exact match with ground-truth.

\textbf{Baselines.} We consider five baselines. \ding{182} \textsc{SFT} that performs the standard supervised fine-tuning on the given task dataset \e{\mathcal{D}_{\mathcal{T}}}. \ding{183} \textsc{Popular} \citep{ghosal2024understandingfinetuningfactualknowledge} that only performs SFT on a subset of \e{\mathcal{D}_{\mathcal{T}}} which the model has knowledge of. This subset is selected either by checking the base model's accuracy on the questions or the monthly views of the entity's Wikipedia page. \ding{184} \textsc{Flame} \citep{lin2024flamefactualityawarealignmentlarge}, which prompts the base model \e{\bm \theta_0} to generate summaries using in-context learning and uses these self-generated summaries as ground-truths for SFT. \ding{185} \textsc{FactTune} \citep{tian2024finetuning} that performs DPO \citep{rafailov2023direct} on top of \ding{182}, where the preference pairs are collected from the base model's sampled outputs (generated by in-context learning as in \ding{184}) and annotated with FActScore. \ding{186} \textsc{RL} \citep{kang2024unfamiliarfinetuningexamplescontrol}, which runs PPO \citep{schulman2017proximalpolicyoptimizationalgorithms} on top of \ding{182}, and the reward model is trained on FActScore annotations of the self-generated summaries in \ding{184}.

\textbf{Implementation details.} We use Llama-3-8B \citep{dubey2024llama3herdmodels} as the base LLM and LoRA to fine-tune all baselines and our method. We search training steps, learning rate, and LoRA rank on the validation set for all methods. For long-form generation, we use multi-version \algname on completely fictitious data, whereas for short QA, we use the basic version on a mix of fictitious and real data. For a fair comparison, all methods access the same real task dataset \e{\mathcal{D}_\mathcal{T}}. For baselines \ding{184}-\ding{186}, we further ensure the total number of training examples matches our method (\emph{i.e.,} our fictitious data contains the same number of entities as \e{\mathcal{D}_\mathcal{T}}, with the same number of responses per entity).

\begin{table}[t]
\caption{Performance for long-form generation tasks (persons and medical entities) and short QA.\quad
\small
$^*$: trained with the same hyperparameters as our method to show the impact of prerequisite learning. $^\dagger$: numbers different from \citet{ghosal2024understandingfinetuningfactualknowledge} because we process ambiguous questions; see Appendix \ref{append:popqa} for results on the original data. $^\ddag$: lower than the original paper because the original model only generates 2.7 claims.}

\small
\label{tab:main-result}
\begin{center}
\vspace{-0.1in}
\begin{tabular}{llllll}
     \toprule \midrule
     & \multicolumn{2}{c}{\textbf{Persons}} & \multicolumn{2}{c}{\textbf{Medical Entities}} & \multicolumn{1}{c}{\textbf{QA}} \\
     & Acc. $\uparrow$     & \# Claims    & Acc. $\uparrow$        & \# Claims   & Acc. $\uparrow$  \\
     \midrule
SFT  & 32.70 & 20.8 & 69.94 & 9.2 & 46.42$^\dagger$ \\
\textsc{Popular} \citep{ghosal2024understandingfinetuningfactualknowledge}  & 41.16 & 15.4 & 65.92 & 8.1 & 45.31 \\
\textsc{Flame} \citep{lin2024flamefactualityawarealignmentlarge}  & 30.32 & 18.2 & 67.92 & 9.8 & -- \\
\textsc{FactTune} \citep{tian2024finetuning}  & 31.93 & 19.6 & 69.13 & 7.9 & -- \\
\textsc{RL} \citep{kang2024unfamiliarfinetuningexamplescontrol} & 33.20$^\ddag$ & 20.9 & 70.03 & 9.0 & -- \\
\midrule
SFT$^{\text{GPT}}$  & 34.75 & 19.7 & 67.98 & 9.0 & -- \\
SFT$^{\text{fictitious}}$$^*$  & 15.44 & 20.6 & 64.44 & 8.9 & 44.98 \\
\rowcolor{gray!20}
\algname & \textbf{45.30} & 16.0 & \textbf{74.35} & 9.1 & \textbf{47.91} \\
\midrule \bottomrule
\end{tabular}
\end{center}
\vspace{-0.15in}
\end{table}

\begin{wrapfigure}{h}{0.4\textwidth}
    \begin{center}
    \vspace{-0.4in}
    \includegraphics[width=\linewidth]{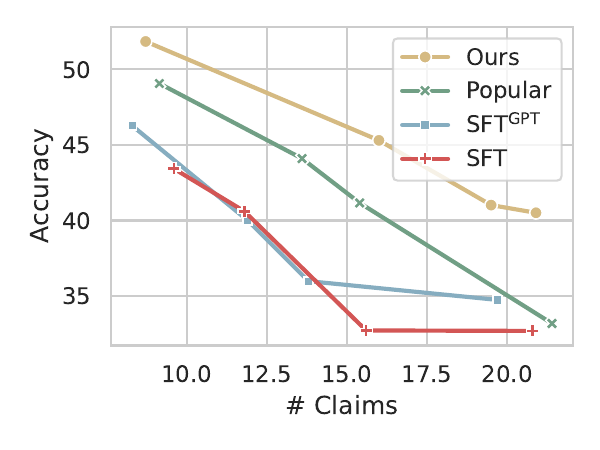}
    \vspace{-0.3in}
    \caption{Accuracy on biography generation under different numbers of generated claims.}
    \label{fig:num-claim}
    \vspace{-0.2in}
    \end{center}
\end{wrapfigure}

\subsection{Main Results}
\label{subsec:main-result}

Table \ref{tab:main-result} presents the main results, where \algname achieves the best performance across all three datasets, especially in long-form generation tasks. It outperforms baselines like \textsc{FactTune} and \textsc{RL}, which use the evaluation metric during training. This suggests the benefits of \algname, where explicitly training for alignment between the model's outputs and internal knowledge provides more direct signals for learning groundedness.

Notably, recall that for long-form generation, \algname was trained on completely fictitious data, and yet it still outperforms all baselines trained on real data. This shows the strong disentanglement power of \algname. To further verify this, we introduce a variant approach, called SFT$^{\text{fictitious}}$, which is also trained on the same fictitious data, but with the prerequisite learning stage removed. The corresponding results in Table \ref{tab:main-result} show the worst performance, which indicates that it is the prerequisite learning that turns the harmful fictitious data into a panacea.

One caveat is that the fictitious data was created using GPT-4, so it is possible that \algname's performance advantage is due to distilling the stronger LLM. To rule out this possibility, we introduce another baseline, SFT$^\text{GPT}$, which is fine-tuned on GPT-4 generated paraphrases of the ground-truth responses in the given dataset \e{\mathcal{D}_{\mathcal{T}}}. The results in Table \ref{tab:main-result} show a significant advantage of \algname over SFT$^\text{GPT}$, which confirms that it is not the distillation of GPT-4, but the disentanglement and groundedness designs that contribute to the superior performance of \algname. Additionally, we also evaluate \algname when GPT-4 is replaced with Llama when constructing the synthetic data. Results in Appendix \ref{append:llama} show similar performance, demonstrating that our method is applicable using only open-source LLMs.

Our initial explorations reveal a negative correlation between accuracy and the number of generated claims. To control this interference, we perform another experiment on the biography generation task where we vary the length of ground-truth responses in the training set to get models that generate different numbers of claims. Results in Figure \ref{fig:num-claim} show that our method consistently achieves the best performance across various numbers of generated claims. Table \ref{tab:sample_outputs} shows some sample outputs.

\subsection{The Knowledge Grounding of \algname}
\label{subsec:rag}
\begin{table}[t]
\centering
\resizebox{\textwidth}{!}{
\begin{minipage}[t]{0.5\textwidth}
\centering
\caption{QA accuracy on fictitious synthetic \textbf{test} data, which contains unseen questions for the skill LoRA \e{\Delta \bm \theta_\textrm{skill}}. Accuracy is computed for two different answers to the same question (V1 and V2).}
\label{tab:RAG-result}
\vspace{-0.01in}
\resizebox{0.9\textwidth}{!}{
\begin{tabular}{lcc}
\toprule \midrule
& \textbf{Acc. V1} & \textbf{Acc. V2} \\ 
\midrule
\e{\bm \theta_0 + \Delta \bm \theta_\textrm{know}^{(1)} + \Delta \bm \theta_\textrm{skill}} & 94.83 & 6.90 \\
\e{\bm \theta_0 + \Delta \bm \theta_\textrm{know}^{(2)} + \Delta \bm \theta_\textrm{skill}} & 13.22 & 95.40 \\
\e{\bm \theta_0 + \Delta \bm \theta_\textrm{skill}} & 15.52 & 5.17 \\
\midrule
SFT$^{\text{real}}$ & 14.94 & 5.17 \\

\midrule \bottomrule
\end{tabular}  
}
\end{minipage}

\hspace{0.01\textwidth}
\begin{minipage}[t]{0.5\textwidth}
\centering
\caption{Performance on fictitious synthetic \textbf{training} data. Memorized Entities measures the percentage of named entities in the fictitious persons' biographies that are memorized.}
\label{tab:pollution-result}
\vspace{-0.01in}
\resizebox{0.9\textwidth}{!}{
\begin{tabular}{lcc}
\toprule \midrule
& \textbf{QA} & \textbf{Bio Generation} \\
& Acc. & Memorized Entities \\ 
\midrule
SFT$^{\text{fictitious}}$ & 58.01 & 32.63$\%$ \\
\e{\bm \theta_0 + \Delta \bm \theta_\textrm{skill}} & 3.99 & 10.79$\%$ \\
\midrule
SFT$^{\text{real}}$ & 3.93 & 10.28$\%$ \\

\midrule \bottomrule

\end{tabular}
}
\end{minipage}
}
\vspace{-0.1in}
\end{table}

Although Section~\ref{subsec:main-result} shows the superior performance of \algname in hallucination reduction, it is still unclear whether \algname truly learns to ground the response to the knowledge in knowledge LoRA as designed. To verify this, we design a knowledge grounding test on the QA task, where we fix the skill LoRA as the one trained for the QA task in the main results, but create two test knowledge LoRAs, \e{\Delta \bm \theta_\textrm{know}^{(1)}} and \e{\Delta \bm \theta_\textrm{know}^{(2)}}. The two test knowledge LoRAs learn two conflicting versions of knowledge about the same fictitious entities. For example, both knowledge LoRAs learn about the fictitious person \emph{Mira Telka}, but version 1 says they are an astrophysicist, and version 2 says they are an architect. It is worth emphasizing that the skill LoRA has \emph{never} seen these test entities, nor the two test knowledge LoRAs, during training.

We then ask the LLM questions about these test fictitious entities with or without the test knowledge LoRAs. Table~\ref{tab:RAG-result} shows the accuracy of LLM's response evaluated against the two conflicting versions of ground-truth answers (Acc. V1 and Acc. V2). As can be observed, when one of the two knowledge LoRAs is plugged in (first two rows), LLM is able to generate the correct answer matching the corresponding knowledge LoRA over 90\% of the time. When the knowledge LoRA is removed (third row), the LLM is unable to answer either version, whose accuracy is as low as the SFT baseline (last row) which has never seen any fictitious data. These results are clear evidence that the LLM answers questions faithfully based on the knowledge provided by the knowledge LoRA.

We find these results very interesting, because they inform an innovative modular design of LLM, where the knowledge LoRA serves as a plug-and-play ``knowledge flash drive'' and the skill LoRA retrieves the knowledge to form an answer. It opens up the possibility of a novel retrieval augmented generation (RAG) paradigm, where the retrieval source is not the external documents, but the knowledge LoRAs, which may address the inference cost and context length challenges.

\subsection{Knowledge Pollution}
\label{subsec:disentanglement}

By design, the skill LoRA does not need to learn any knowledge information because the knowledge LoRA already covers it all. However, since the skill LoRA is exposed to the knowledge information in the task dataset \e{\mathcal{D}_\mathcal{T}} during training, one might wonder whether any knowledge could accidentally get picked up by the skill LoRA, a phenomenon we refer to as \emph{knowledge pollution}. Knowledge pollution, if present, would indicate that the disentanglement of \algname is incomplete, and undermine the reliability of the skill LoRA, especially when the new knowledge is fictitious.

To test whether knowledge pollution is present, we plug the skill LoRA from the main results into the LLM, remove the knowledge LoRA, and ask LLM questions about the training knowledge. Note that the main difference from Section~\ref{subsec:rag} is that here the questions are about fictitious training knowledge the skill LoRA has seen, whereas Section~\ref{subsec:rag} is about unseen test knowledge. Table~\ref{tab:pollution-result} shows the QA accuracy and the percentage of memorized entities for biography generation (the latter is computed as the percentage of named entities in the fictitious persons' knowledge that are generated by the model). The results indicate that without any knowledge LoRAs, the skill LoRA alone (second row) is unable to answer the questions correctly, and its behavior is similar to SFT$^\text{real}$, which was trained on real data only and thus guaranteed no knowledge pollution. As a reference, SFT$^\text{fictitious}$, which is trained directly on the fictitious knowledge, achieves a much higher performance. 

These results show no evidence of knowledge pollution, and thus verify the strong disentanglement power of \algname. It also inspires potential privacy protection applications, where the skill LoRA can be trained on private data to learn the skills without memorizing any protected knowledge.

\subsection{Scaling Synthetic Data}
\label{subsec:scale}
As discussed, synthetic data has the benefit of cheap scaling. We are thus interested in what would happen after such scaling. Figure~\ref{fig:num-data} shows the QA and biography generation performance (detailed settings in Appendix \ref{append:scale}) as the amount of fictitious synthetic data increases. The blue line shows the performance of \algname, which shows an increasing trend. The red line shows the performance of directly fine-tuning on the fictitious synthetic data without prerequisite learning, which exhibits clear degradation. This intriguing contrast confirms that while the fictitious synthetic data is inherently harmful, \algname endows it with a positive scaling effect.

\begin{figure}
    \centering
    \begin{subfigure}{0.45\textwidth}
        \centering
        \includegraphics[width=\linewidth]{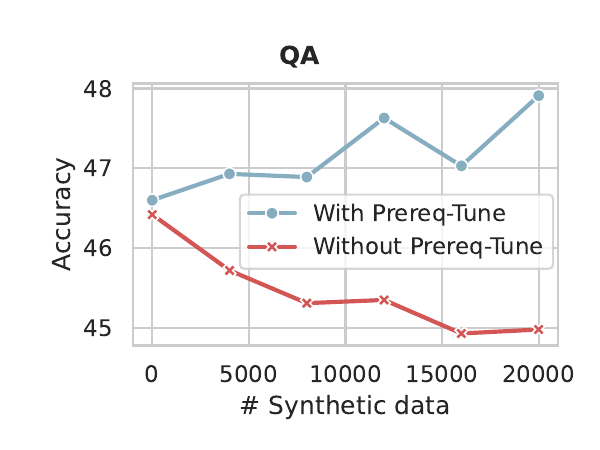}
    \end{subfigure}
    \hspace{0.01\textwidth}
    \begin{subfigure}{0.45\textwidth}
        \centering
        \includegraphics[width=\linewidth]{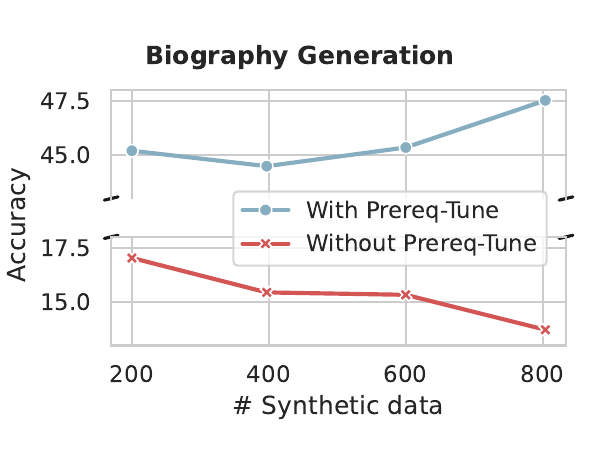}
    \end{subfigure}
    \vspace{-0.15in}
    \caption{Performance as the number of synthetic data scales up.}
    \label{fig:num-data}
    \vspace{-0.1in}
\end{figure}

\subsection{Answer Abstention and Verbalized Uncertainty}
\label{subsec:abstention_exper}

\begin{wrapfigure}{h}{0.4\textwidth}
    \begin{center}
    \vspace{-0.2in}
    \includegraphics[width=\linewidth]{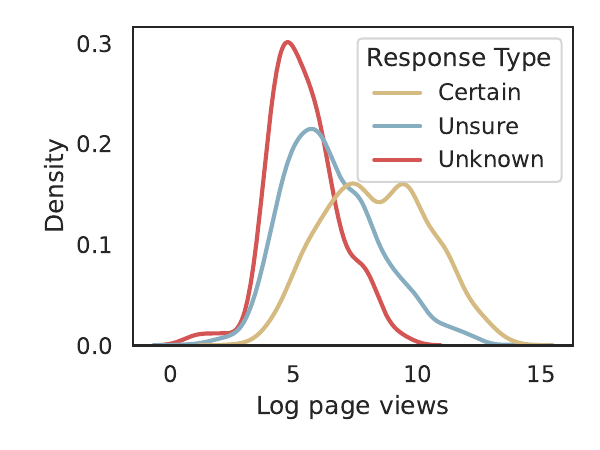}
    \vspace{-0.2in}
    \caption{The distribution of each response type with respect to the log of monthly page views of the entities.}
    \label{fig:uncertainty}
    \vspace{-0.1in}
    \end{center}
\end{wrapfigure}

Finally, we evaluate the performance of the extended version of \algname that also teaches the LLM to say \emph{`I don't know'} (Section \ref{subsec:abstention}). Specifically, for the QA task,  we create multi-version dataset pairs \e{(\mathcal{D}_{\textrm{know}}^{(k)}, \mathcal{D}_{\mathcal{T}}^{(k)})}, such that for each question, one of the pairs turns the corresponding knowledge into unfamiliar knowledge (with number of paraphrases \e{M=0}), and the corresponding answer into \emph{`I don't know'} (IDK). We compare with a baseline that splits the real data into unknown and unknown parts and trains the model for IDK responses on unknown data \citep{zhang-etal-2024-r}. On the test set, for questions that the model knows, our method improves correctly answered questions from \e{53.20\%} to \e{54.94\%} and reduces mistakenly abstained questions from \e{35.51\%} to \e{33.64\%}. For unknown questions, our method increases the IDK responses from \e{85.32\%} to \e{85.63\%}.

The results above confirm that the number of knowledge paraphrases \e{M} can effectively control the familiarity of the knowledge, which inspires us to use \e{M} to enable a more fine-grained expression for uncertainty. Specifically, in addition to \textit{certain} and \textit{unknown}, we introduce a third level of familiarity, \textit{unsure}, in between by setting another threshold on \e{M}. The answers involving the unsure knowledge are rewritten as \textit{`I think it might be ...'} (See Appendix~\ref{append:uncertainty_data} for more details). 
Figure \ref{fig:uncertainty} shows the distribution of the three response types to questions on real entities, across the log of monthly views of the entity's Wikipedia page, which is a rough estimation of the entity's familiarity. As shown in the figure, the model's response type generally aligns with the entity's familiarity, indicating that the skill of verbalizing uncertainty, which is learned only on fictitious synthetic data, can generalize reasonably to the base model's pre-trained knowledge.

\section{Conclusion}
In this paper, we propose \algname, a novel fine-tuning strategy to reduce LLM hallucinations. \algname disentangles the learning of skills and knowledge by introducing a prerequisite learning stage, which equips LLMs with the necessary knowledge required for subsequent SFT. Moreover, \algname can be combined with fictitious synthetic data to improve the groundedness of LLMs' generations. Experiments on three datasets show that \algname outperforms strong baselines in reducing hallucinations. Further analyses also verify its disentanglement.

\section{Acknowledgements}
The work of Yujian Liu and Shiyu Chang was partially supported by National Science Foundation (NSF) Grant IIS-2338252, NSF Grant IIS-2207052, and NSF Grant IIS-2302730. The computing resources used in this work were partially supported by the Accelerate Foundation Models Research program of Microsoft. Tommi Jaakkola acknowledges support from the MIT-IBM Watson AI Lab and the NSF Expeditions grant (award 1918839) Understanding the World Through Code.

\section{Ethics Statement}
Our work aims to enhance the reliability and trustworthiness of LLMs by reducing the hallucinations in their generations. However, while our method shows clear improvements over baselines, it does not eliminate hallucinations. In fact, results in Table \ref{tab:main-result} and sample outputs in Table \ref{tab:sample_outputs} show that the fine-tuned model can still generate factually incorrect statements. Therefore, users should remain cautious when using our fine-tuned model and are strongly advised to verify its outputs through additional, trusted information sources. Furthermore, although our analyses in Section \ref{subsec:disentanglement} show promising results for privacy protection, care should be taken to ensure that no sensitive information is inadvertently included in real-world deployments.
\section{Reproducibility Statement}
We have taken the necessary steps to ensure the reproducibility of our results. Specifically, Section \ref{subsec:exp-setting} discusses the general experiment settings in our paper. Appendix \ref{append:real-data} provides the detailed steps to collect and process the datasets used in downstream tasks. Appendix \ref{append:prereq-data} includes the detailed steps to construct the fictitious synthetic data used by our method. Finally, Appendix \ref{append:implementation} and the supplementary material list the implementation details of our method and all baselines, including the codebase, training hyperparameters, evaluation details, \emph{etc.}

\newpage

\bibliography{iclr2025_conference}
\bibliographystyle{iclr2025_conference}

\newpage

\appendix
\makeatletter
\lst@InstallKeywords k{attributes}{attributestyle}\slshape{attributestyle}{}ld
\makeatother

\lstdefinestyle{courierStyle}{
    basicstyle=\fontsize{8}{9}\fontfamily{pcr}\selectfont,
    showstringspaces=false,
    breaklines=true,
    breakatwhitespace=false,
    breakindent=0pt,
    keepspaces=false,
    showspaces=false,   
    escapeinside={(*@}{@*)}
}

\section{Task Datasets}
\label{append:real-data}
This section presents the details of collecting and processing the real task dataset \e{\mathcal{D}_\mathcal{T}} for three tasks in our experiments. Table \ref{tab:real_task_data} shows the statistics for three datasets.

\begin{table}[t]
\centering
\caption{Number of examples in the real downstream task dataset \e{\mathcal{D}_\mathcal{T}}.}
\label{tab:real_task_data}
\resizebox{0.6\textwidth}{!}{
\begin{tabular}{lccc}
\toprule \midrule
 & \textbf{Persons} & \textbf{Medical Entities} & \textbf{Short QA} \\
 \midrule
Training & 397 & 449 & 10,613 \\
Validation & 60 & 80 & 789 \\
Test & 183 & 200 & 2,152 \\
 \midrule \bottomrule
\end{tabular}
}
\end{table}

\subsection{Long-form Generation}
For long-form generation tasks, we collect real-world entities from Wikidata and use the first section of their Wikipedia page as the ground-truth response. Specifically, we collect entities within the category of ``Human'' for biography generation, and entities in ``class of disease'' and ``medicinal product'' for medical QA. Following prior works \citep{min-etal-2023-factscore}, we consider entities whose Wikipedia page's monthly views are greater than 1,000 as popular. For biography generation, we format the instruction-response pair as `\textit{Question: Generate a biography for \{entity\}. Answer: \{response\}}'. For medical QA, the format is `\textit{Question: Tell me about \{entity\}. Answer: \{response\}}'.

\subsection{Short QA}
For the short QA task, we evaluate on PopQA \citep{mallen-etal-2023-trust} and randomly split the data into training, validation, and test to ensure no overlapping subjects. During the preliminary study, we find PopQA contains many ambiguous questions due to the ambiguity in the asked subject (\emph{e.g.,} the question `\textit{Who was the director of Legacy?}' could refer to multiple films with the same name \textit{Legacy}). We thus clarify these questions by replacing the subject in the question with its Wikipedia title, which links to a unique real-world entity. The results on the original dataset are also reported in Section \ref{append:popqa}. We format instruction-response pair as `\textit{Question: \{question\}. Answer: \{answer\}}' for training.

\section{Dataset Pairs for \algname}
\label{append:prereq-data}

We now describe the detailed procedure for constructing the dataset pair \e{(\mathcal{D}_\textrm{know}, \mathcal{D}_\mathcal{T})} for \algname on each task. Table \ref{tab:synthetic_data} summarizes the statistics of the constructed datasets.

\subsection{Basic \algname for Short QA}
\label{append:qa_data}
For the short QA task, we use a combination of fictitious and real data. For real data, we construct \e{\mathcal{D}_\textrm{know}} from existing \e{\mathcal{D}_\mathcal{T}} using the top-down strategy. Specifically, since questions in PopQA are created from knowledge triplets (subject, relation, object) using templates, we prompt GPT-4 to convert each QA template into a statement. For example, the QA pair `\textit{Question: Who was the director of \{subject\}?} \textit{Answer: \{object\}}' can be converted to a statement with `\textit{\{subject\} was directed by \{object\}.}' This template is further paraphrased 15 times by GPT-4, and the statement-based knowledge dataset is constructed by replacing the subject-object pairs in \e{\mathcal{D}_\mathcal{T}} into these templates. For the passage-based knowledge dataset, we directly use the Wikipedia page of the subjects.

For fictitious data, we create \e{\mathcal{D}_\textrm{know}} first and then convert it into the task dataset \e{\mathcal{D}_\mathcal{T}} using the bottom-up strategy. Specifically, we prompt GPT-4 to generate triplets (subject, relation, object) where the subject is a fictitious entity (detailed prompt in Figure \ref{fig:qa-fictitious-prompt}). We then construct the knowledge dataset following the above procedure. To create the task dataset, we replace these fictitious triplets into the original QA templates used in PopQA.

\begin{table}[t]
\centering
\caption{Statistics of the synthetic datasets for \algname.}
\label{tab:synthetic_data}
\resizebox{0.75\textwidth}{!}{
\begin{tabular}{lccc}
\toprule \midrule
 & \textbf{Persons} & \textbf{Medical Entities} & \textbf{Short QA} \\
 \midrule
\# entities & 397 & 449 & 20,000 \\
\# knowledge versions & 5 & 5 & 1 \\
\# sentences per version & 6.5 & 4.6 & 1 \\
 \midrule \bottomrule
\end{tabular}
}
\end{table}

\subsection{Multi-version \algname for Long-form Generation}
For long-form generation tasks about persons and medical entities, we use completely fictitious data for multi-version \algname. The dataset construction combines both top-down and bottom-up strategies. In the following, we describe the detailed steps for biography generation. Datasets for medical entities are created similarly using the same prompts, except that mentions of persons are changed to medical entities.

\textbf{Step 1: Construct seed fictitious biographies.} We prompt GPT-4 to generate a biography for a fictitious person, using real persons' Wikipedia pages as references. The detailed prompt is listed in Figure \ref{fig:bio-fictitious-prompt}. We further filter out fictitious persons whose names coincide with a real Wikipedia entry. The remaining biographies are called \textit{seed biographies}.

\textbf{Step 2: Get individual statements.} We decompose the seed biographies into individual statements. We experiment with two decomposition methods. First, we prompt GPT-4 to break down the biographies into atomic claims (instruction in Figure \ref{fig:decompose-prompt}). Second, we break down the biographies into sentences and consider each sentence as an individual statement. The experiments in the main results adopt the first method, and Appendix \ref{append:ablation} further compares these two methods.

\textbf{Step 3: Construct multi-version knowledge dataset.} To create multiple versions of the knowledge dataset, we randomly sample the statements of a person into different subsets and create a version of knowledge for each subset. Particularly, we employ a two-step sampling procedure. First, we uniformly sample the number of statements to cover, denoted as \e{n}. Then we sample \e{n} statements from the total set of statements without replacement. The resulting statements are each paraphrased 15 times to construct one version of the statement-based knowledge for the person. To create passage-based knowledge, we prompt GPT-4 to generate a summary for the person based on the subset of statements. The detailed prompt is listed in Figure \ref{fig:bio-task-prompt}. Each summary is then paraphrased 5 times to construct one version of the passage-based knowledge for the person.

\textbf{Step 4: Construct multi-version task dataset.} The task dataset is created from the individual statements using the bottom-up strategy. Specifically, for each subset of statements, we use the prompt in Figure \ref{fig:bio-task-prompt} to instruct GPT-4 to generate a biography, which is considered as the ground-truth response for the corresponding version of knowledge.

\subsection{Multi-version \algname for Verbalized Uncertainty}
\label{append:uncertainty_data}
We use completely fictitious data for the verbalized uncertainty experiment. We modify the fictitious data in Section \ref{append:qa_data} to create the multi-version dataset pairs. Specifically, for each knowledge piece in \e{\mathcal{D}_\textrm{know}}, we create three versions of statement-based knowledge with different numbers of paraphrases. In \e{\mathcal{D}_\textrm{know}^{(1)}}, each statement is paraphrased 15 times. This corresponds to the most familiar version of the knowledge. In \e{\mathcal{D}_\textrm{know}^{(2)}} and \e{\mathcal{D}_\textrm{know}^{(3)}}, each statement is paraphrased 6 and 1 times respectively, which correspond to the moderate and least familiar versions. We then modify the task dataset so that the ground-truth response aligns with the model's familiarity. Particularly, in \e{\mathcal{D}_\mathcal{T}^{(1)}}, we use the answer template `\textit{I'm sure the answer is ...}'. In \e{\mathcal{D}_\mathcal{T}^{(2)}} and \e{\mathcal{D}_\mathcal{T}^{(3)}}, we use the templates `\textit{I think it might be ...}' and `\textit{I don't know, my guess is ...}'.

\section{Implementation Details}
\label{append:implementation}

\begin{table}[t]
\centering
\caption{Training hyperparameters in our experiments.}
\label{tab:hyperparameter}
\resizebox{0.75\textwidth}{!}{
\begin{tabular}{lccc}
\toprule \midrule
 & \textbf{Persons} & \textbf{Medical Entities} & \textbf{Short QA} \\
 \midrule
Epochs & \e{5, 10, 20, 30, \ldots, 80} & \e{5, 10, 20, 30, \ldots, 80} & \e{3,4,5} \\
learning rate & \e{3e-5}, \e{5e-5} & \e{3e-5}, \e{5e-5} & \e{3e-5}, \e{5e-5} \\
Batch size & \e{128} & \e{128} & \e{512} \\
LoRA \e{r} & \e{32, 64, 128} & \e{32, 64, 128} & \e{16, 32, 64} \\
LoRA \e{\alpha} & \e{2*r} & \e{2*r} & \e{2*r} \\
\midrule \bottomrule
\end{tabular}
}
\end{table}

We base our implementations on alignment-handbook.\footnote{\url{https://github.com/huggingface/alignment-handbook}.} All experiments are conducted on 4 80GB NVIDIA A100 GPUs. During inference, we use greedy decoding for all methods. We tune the number of training epochs, learning rate, and LoRA rank on the validation set for all methods. Table \ref{tab:hyperparameter} lists the hyperparameters we search.

On long-form generation tasks, we observe that models sometimes generate low-quality responses with many repetitions. We thus filter out checkpoints when more than \e{10\%} of their generations have a sep-rep-4 score higher than 0.2, as calculated in \citet{Welleck2020Neural}.

To evaluate generated summaries on long-form generation tasks, we use FActScore \citep{min-etal-2023-factscore} with GPT-4o-mini as the underlying LLM. To prevent models from exploiting the metric by generating generic or subjective statements, we modify the evaluation pipeline and add a filtering step after decomposing the generated summary, so that only statements presenting objective and concrete information are evaluated. Figure \ref{fig:filter-prompt} lists the prompt we use for this filtering.

\subsection{Baselines}
For \textsc{Popular} \citep{ghosal2024understandingfinetuningfactualknowledge}, on long-form generation tasks, we train on the subset of entities whose monthly Wikipedia page views are greater than 1,000. On the short QA task, we follow the definition in \citet{gekhman2024doesfinetuningllmsnew} to split the training set into known and unknown questions and only fine-tune on known questions.

For \textsc{Flame} \citep{lin2024flamefactualityawarealignmentlarge}, we follow settings in the original paper to sample outputs from the base model \e{\bm \theta_0} using in-context learning with 5 demonstrations. We randomly sample training examples in \e{\mathcal{D}_\mathcal{T}} to serve as demonstrations. 5 responses are sampled for each entity to match the amount of data used by our method. To ensure the high quality of training examples, we filter out samples with a sep-rep-4 score higher than 0.2.

For \textsc{FactTune} \citep{tian2024finetuning}, we evaluate the 5 responses above using FActScore and collect \e{\binom{5}{2}} preference pairs for each entity, where the response with a higher score is considered as the chosen response and the other one as the rejected response. We set \e{\beta=0.1}, learning rate as \e{1e-6}, and train for 500 steps following \citep{lin2024flamefactualityawarealignmentlarge}.

For \textsc{RL} \citep{kang2024unfamiliarfinetuningexamplescontrol}, we use the official implementation and train the reward model using the above FActScore annotations.

\subsection{Basic \algname for Short QA}
On short QA, we use a combination of fictitious and real data. During prerequisite learning, we train separate knowledge LoRAs for fictitious and real entities. For both fictitious and real entities, we further train separate knowledge LoRAs on statement-based and passage-based knowledge.

During SFT, we randomly sample a knowledge LoRA at each iteration, either statement-based or passage-based, and train the skill LoRA on top of it. Additionally, to mitigate the training and inference gap, we add a regularization term that trains the skill LoRA on top of the base model, without knowledge LoRA, which leads to the following optimization problem:
\begin{equation}
    \small
    \min_{\Delta \bm \theta_\textrm{skill}} \mathcal{L}_\mathcal{T}(\bm \theta_0 + \Delta \bm \theta_\textrm{know} + \Delta \bm \theta_\textrm{skill}; \mathcal{D}_{\mathcal{T}}) + \alpha \mathcal{L}_\mathcal{T}(\bm \theta_0 + \Delta \bm \theta_\textrm{skill}; \mathcal{D}_{\mathcal{T}}^{\text{real}}),
\end{equation}
where \e{\mathcal{D}_{\mathcal{T}}^{\text{real}}} is the subset in \e{\mathcal{D}_{\mathcal{T}}} that contains only real entities, and \e{\alpha} is a hyperparameter.

\subsection{Multi-version \algname for Long-form Generation}

On long-form generation tasks, we use completely fictitious data. For each knowledge dataset \e{\mathcal{D}_{\mathrm{know}}^{(k)}}, we train two knowledge LoRAs, one for statement-based and one for passage-based.

During SFT, at each iteration, we first uniformly sample a version from all possible versions, \e{k \sim \mathcal{U}(\{1,\ldots,K\})}. Then we randomly select either the statement-based or passage-based knowledge LoRA for this version, and train the skill LoRA on top of it using the task dataset \e{\mathcal{D}_{\mathcal{T}}^{(k)}}.

\section{Additional Results}

\subsection{Performance of \algname Using Llama}
\label{append:llama}

\begin{wrapfigure}{h}{0.45\textwidth}
    \begin{center}
    \vspace{-0.35in}
    \includegraphics[width=\linewidth]{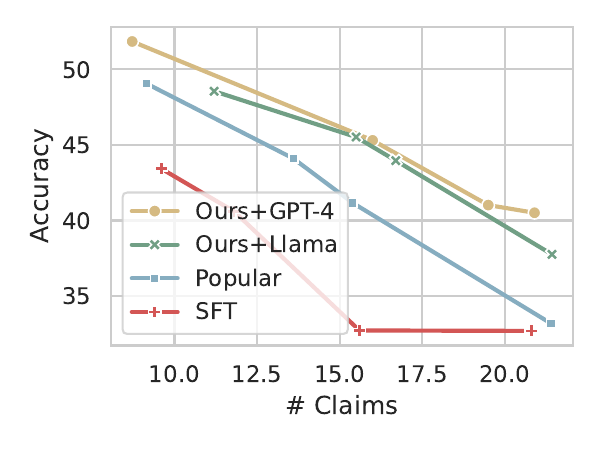}
    \vspace{-0.3in}
    \caption{Accuracy on biography generation under different numbers of generated claims.}
    \label{fig:num-claim-llama}
    \vspace{-0.3in}
    \end{center}
\end{wrapfigure}

We further evaluate \algname when Llama-3.1-70B-Instruct \citep{dubey2024llama3herdmodels} is used to construct the fictitious synthetic data instead of GPT-4. Results in Figure \ref{fig:num-claim-llama} show that using Llama achieves similar performance with GPT-4, demonstrating that \algname can be applied when only open-source LLMs are available.

\subsection{Performance on Original PopQA}
\label{append:popqa}

\begin{table}[t]
\centering
\resizebox{0.95\textwidth}{!}{
\begin{minipage}[t]{0.36\textwidth}
\centering
\caption{Accuracy on the original PopQA, without data cleaning.}
\label{tab:qa_original}

\resizebox{0.75\linewidth}{!}{
\begin{tabular}{lc}
\toprule \midrule
& \textbf{Accuracy} \\
\midrule
SFT & 36.90 \\
\textsc{Popular} & 36.85 \\
SFT$^{\text{fictitious}}$ & 36.05 \\
\algname & 37.50 \\
\midrule \bottomrule
\end{tabular}
}
\end{minipage}

\hspace{0.03\textwidth}
\begin{minipage}[t]{0.43\textwidth}
\centering
\caption{Performance of different formats for the knowledge dataset \e{\mathcal{D}_\textrm{know}}.}
\label{tab:knowledge_ablation}

\resizebox{\linewidth}{!}{
\begin{tabular}{lcc}
\toprule \midrule
& \textbf{QA} & \textbf{Bio Generation} \\
& Accuracy & Accuracy \\ 
\midrule
Both & 47.91 & 45.30 \\
Statement-based & 47.58 & 38.75 \\
Passage-based & 47.07 & 39.75 \\
\midrule \bottomrule
\end{tabular}
}
\end{minipage}
}
\end{table}

Table \ref{tab:qa_original} shows the results on the original PopQA dataset, without clarifying questions. Although the numbers are different from those in Table \ref{tab:main-result}, the general trend is similar: \algname achieves the best performance whereas SFT$^{\text{fictitious}}$ is the worse, which again demonstrates the impact of the proposed prerequisite learning stage.

\subsection{Ablation Study}
\label{append:ablation}
We now investigate the influence of two design choices in \algname.

\textbf{Data formats in the knowledge datasets.} Table \ref{tab:knowledge_ablation} shows the performance of only using statement-based, passage-based, or both formats on short QA and biography generation tasks. As can be observed, combining both formats achieves better performance than using any of the formats alone. We hypothesize that knowledge is stored differently in knowledge LoRAs trained on these two formats, so training the skill LoRA to work with both formats improves its generalizability to different knowledge representations.

\begin{wrapfigure}{h}{0.45\textwidth}
    \begin{center}
    \vspace{-0.2in}
    \includegraphics[width=\linewidth]{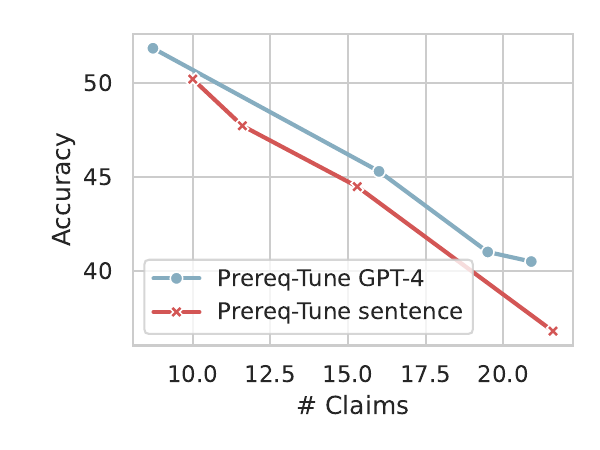}
    \vspace{-0.3in}
    \caption{Accuracy on biography generation with two decomposition methods.}
    \label{fig:bio-decompose}
    \vspace{-0.1in}
    \end{center}
\end{wrapfigure}

\textbf{Methods to decompose seed biographies.} We compare two methods to decompose seed biographies into individual statements. The first method prompts GPT-4 for decomposition (GPT-4-based), and the second method simply considers each sentence as a statement (sentence-based). The main difference between the two is that GPT-4 breaks down biographies into atomic statements that cannot be further decomposed, whereas sentence-based decomposition may contain multiple facts within each statement (\emph{e.g.,} \textit{`Emma Louise Carter (born May 22, 1965) is an American politician, entrepreneur, and philanthropist who served as the 48th vice president of the United States from 2009 to 2017 under President James Matthews.'}). Results in Figure \ref{fig:bio-decompose} show that GPT-4-based decomposition outperforms sentence-based decomposition, which suggests that more fine-grained decomposition benefits the model in learning groundedness.

\subsection{Sample Outputs}
\begin{table}[t]
\centering
\resizebox{\textwidth}{!}{
\begin{tabular}{>{\centering\arraybackslash}m{0.1\textwidth} >{\raggedright\arraybackslash}m{\textwidth}
}
\toprule \midrule 
\rowcolor{transgray} 
\multicolumn{2}{c}{\textbf{Input Instruction}} \\ \midrule
& Generate a biography for Eliza Dushku. \\
\midrule
\rowcolor{transgray} 
\multicolumn{2}{c}{\textbf{Model Responses}} \\ \midrule
\textsc{SFT} & Eliza Dushku (born \highlight[green!20]{December 30, 1980}) is an \highlight[green!20]{American actress and model}. She is known for her role as \highlight[red!20]{Kelly Reynolds} in the cable broadcast series \highlight[green!20]{Dollhouse (2009–2010)}. She also appeared in the films \highlight[green!20]{True Lies (1994)}, \highlight[red!20]{The Tartuffe (2001)}, \highlight[green!20]{Bring It On (2000)}, \highlight[red!20]{Dog Days of Summer (2007)}, \highlight[red!20]{Tortured (2008)}, \highlight[red!20]{Gunshot Straight (2008)}, \highlight[red!20]{Miss Match (2003)}, \highlight[red!20]{Wrong Turn at Tahoe (2009)}, \highlight[green!20]{Eloise (2017)}, and \highlight[red!20]{Jules (2019)}. \highlight[red!20]{From 2014 to 2017, she portrayed Athena on FX's Sons of Anarchy and its spin-off Mayans M.C..} Dushku also performed the \highlight[red!20]{voice of Athena in the video games True Lies (1994) and as Jessica in the first two Guild Wars games.} \\
\midrule
SFT$^{\text{GPT}}$ & Eliza Patrice Dushku (born \highlight[green!20]{December 30, 1980}) is an \highlight[green!20]{American actress and model}. She is recognized for her roles as \highlight[red!20]{FBI Special Agent Kimberly Price in the TNT series Bull}, as \highlight[red!20]{Athena on the Syfy series Dollhouse}, and as the \highlight[green!20]{voice of Catwoman} in the animated film \highlight[red!20]{Batman: Assault on Arkham}. Dushku has also appeared in films such as \highlight[green!20]{True Lies (1994)}, \highlight[red!20]{The Scorpion King (2002)}, \highlight[green!20]{Wrong Turn (2003)}, \highlight[green!20]{Jay and Silent Bob Strike Back (2001)}, \highlight[green!20]{Bring It On (2000)}, and \highlight[red!20]{Soul Man (2014)}. She had recurring roles on the \highlight[green!20]{series Tru Calling} and Angel and has featured in the \highlight[red!20]{video games True Crimes: Streets of L.A. and Dog Day Afternoon.} \\
\midrule
\textsc{Popular} & Eliza Patricia Dushku (born \highlight[green!20]{December 30, 1980}) is an \highlight[green!20]{American actress and model}. She is known for her roles as \highlight[green!20]{Faith in the action film series Buffy the Vampire Slayer} \highlight[red!20]{(1992–2003)} and its spin-off television series \highlight[green!20]{Angel (1999–2004)}, as well as her roles in \highlight[green!20]{True Lies (1994)}, \highlight[green!20]{Bring It On (2000)}, \highlight[green!20]{Jay and Silent Bob Strike Back (2001)}, \highlight[green!20]{Wrong Turn (2003)}, and \highlight[green!20]{Dollhouse (2009–2010)}. She also voiced Catwoman in the \highlight[green!20]{animated film Catwoman} \highlight[red!20]{(2004)} and \highlight[red!20]{voiced the character of Chelsea in the video game Mass Effect 3 (2012).} \\
\midrule
\algname & Eliza Patricia Dushku (born \highlight[green!20]{December 30, 1980}) is an \highlight[green!20]{American actress, model, and activist}. She is best known for her television roles as \highlight[green!20]{Faith on the action series Buffy the Vampire Slayer (1998–2003)} and \highlight[green!20]{its spin-off series Angel (2000–2004)}, and as \highlight[red!20]{Jessica Angel} in \highlight[green!20]{True Lies (1994)}. She also starred in the \highlight[green!20]{Fox series Dollhouse (2009–2010)} and appeared in the films \highlight[green!20]{True Lies (1994)}, \highlight[green!20]{The New Guy (2002)}, \highlight[green!20]{Bring It On (2000)}, \highlight[green!20]{Jay and Silent Bob Strike Back (2001)}, \highlight[green!20]{Soul Survivors (2001)}, \highlight[green!20]{Wrong Turn (2003)}, and \highlight[green!20]{The Alphabet Killer} (\highlight[red!20]{2007}). \\
\midrule
\bottomrule
\end{tabular}
}
\caption{Example input instruction and model responses on the biography generation task. We mark factually correct information in \highlight[green!20]{\textbf{green}} and hallucinated information in \highlight[red!20]{\textbf{red}}.}
\label{tab:sample_outputs}
\end{table}

Table \ref{tab:sample_outputs} shows model responses on an example for the biography generation task, which illustrates that our method generates less hallucinated content.

\subsection{Experiment Settings for Scaling Synthetic Data}
\label{append:scale}

Figure \ref{fig:num-data} shows the performance when we scale up the amount of synthetic data. Specifically, for short QA, we increase the number of synthetic questions. For biography generation, we increase the number of fictitious synthetic persons and use sentence-based decomposition for \algname.

\begin{figure}[h]
\centering
\begin{tcolorbox}
\begin{lstlisting}[style=courierStyle]
Given the following (subject, object) pairs that have the relation of {relation}, generate 10 fictitious pairs that have the same relation.
{example subject-object pairs}

- Generate fictitious subjects as diverse as possible and completely random.
- Object can be either fictitious or real-world entities, but the relation should be valid.
- Output each pair at a line in JSON format with keys "subject" and "object".
- Directly generate results and nothing else.
\end{lstlisting}
\end{tcolorbox}
\caption{Prompt used for creating fictitious data for the QA task.}
\label{fig:qa-fictitious-prompt}
\end{figure}

\begin{figure}[h]
\begin{center}
\begin{tcolorbox}
\begin{lstlisting}[style=courierStyle]
Here is the Wikipedia biography for {real person}:
{wiki summary}

Using this as a reference, generate a biography for a completely fictitious person.

Requirements:
- The person should be fictitious.
- The biography should resemble the provided Wikipedia biography in structure but pertain to the fictitious person.
- Do not mention "fictitious" or any other indication that the person is not real.
- Generate outputs in valid json string format with keys "person" and "biography".
\end{lstlisting}
\end{tcolorbox}
\caption{Prompt used for creating fictitious biographies.}
\label{fig:bio-fictitious-prompt}
\end{center}
\end{figure}

\begin{figure}[h]
\begin{center}
\begin{tcolorbox}
\begin{lstlisting}[style=courierStyle]
Generate a biography for the person {fictitious person} based on the following facts.
Facts:
{set of statements}

Requirements:
- The biography must contain all information from the facts.
- Only include information from the facts. Do not add any other information.
- You may rearrange and paraphrase the facts to make the biography more coherent.
- Directly generate the biography and nothing else.
\end{lstlisting}
\end{tcolorbox}
\caption{Prompt used for creating a biography from a set of statements.}
\label{fig:bio-task-prompt}
\end{center}
\end{figure}

\begin{figure}[h]
\begin{center}
\begin{tcolorbox}
\begin{lstlisting}[style=courierStyle]
Given the following claims about {entity}, please select the ones that present concrete and specific information. Omit any claims that are generic or purely subjective.

Examples of claims with concrete information:
{positive_examples}

Examples of generic or subjective claims:
{negative_examples}

Claims:
{decomposed claims}

Requirements:
- Copy each concrete claim after the "-" and nothing else.
\end{lstlisting}
\end{tcolorbox}
\caption{Prompt used for filtering out generic or subjective claims in FActScore.}
\label{fig:filter-prompt}
\end{center}
\end{figure}

\begin{figure}[h]
\begin{center}
\begin{tcolorbox}
\begin{lstlisting}[style=courierStyle]
Please break down the following biography about {person} into independent facts. Focus on facts that present concrete and objective information.
Biography:
{biography}

Requirements:
- Generate atomic facts that cannot be further decomposed.
- Include all information from the biography in the facts.
- Facts should target {person}, not other entities.
- Output each fact at a separate line after "- " and nothing else.
\end{lstlisting}
\end{tcolorbox}
\caption{Prompt used for decomposing seed biographies into individual statements.}
\label{fig:decompose-prompt}
\end{center}
\end{figure}

\end{document}